\documentclass{article}

\usepackage{microtype}
\usepackage{graphicx}
\usepackage{subfigure}
\usepackage{booktabs} % for professional tables

\usepackage{hyperref}

\usepackage[accepted]{icml2025}

\usepackage{amsmath}
\usepackage{amssymb}
\usepackage{mathtools}
\usepackage{amsthm}

\usepackage{tabularx}
\usepackage{graphicx}
\usepackage{algorithm}
\usepackage{algorithmic}
\usepackage{booktabs}
\usepackage{multirow}
\usepackage{enumitem}
\usepackage{tikz}
\usepackage{wrapfig}
\usetikzlibrary{positioning, arrows}

\usepackage[capitalize,noabbrev]{cleveref}

\theoremstyle{plain}

\theoremstyle{definition}

\theoremstyle{remark}

\usepackage[textsize=tiny]{todonotes}

\icmltitlerunning{PAACE: A Plan-Aware Automated Agent Context Engineering Framework}

\begin{document}

\twocolumn[
\icmltitle{{PAACE: A Plan-Aware Automated Agent Context Engineering Framework}}




\begin{icmlauthorlist}
\icmlauthor{Kamer Ali Yuksel}{comp}
\end{icmlauthorlist}

\icmlaffiliation{comp}{aiXplain Inc, San Jose, USA}

\icmlcorrespondingauthor{Kamer Ali Yuksel}{kamer@aixplain.com}
\icmlkeywords{Machine Learning, ICML}

\vskip 0.3in
]



\printAffiliationsAndNotice{}  % leave blank if no need to mention equal contribution

\begin{abstract}
Large Language Model (LLM) agents are increasingly deployed in complex, multi-step workflows involving planning, tool use, reflection, and interaction with external knowledge systems. These workflows generate rapidly expanding contexts that must be curated, transformed, and compressed to maintain fidelity, avoid attention dilution, and reduce inference cost. Prior work on summarization and query-aware compression largely ignores the \emph{multi-step, plan-aware} nature of agentic reasoning. In this work, we introduce \textbf{PAACE} (\textbf{P}lan-\textbf{A}ware \textbf{A}utomated \textbf{C}ontext \textbf{E}ngineering), a unified framework for optimizing the evolving state of LLM agents through next-$k$-task relevance modeling, plan-structure analysis, instruction co-refinement, and function-preserving compression. PAACE comprises (1) \textbf{PAACE-Syn}, a large-scale generator of synthetic agent workflows annotated with stepwise compression supervision, and (2) \textbf{PAACE-FT}, a family of distilled, plan-aware compressors trained from successful teacher demonstrations. Experiments on long-horizon benchmarks---AppWorld, OfficeBench, and 8-Objective QA---demonstrate that PAACE consistently improves agent correctness while substantially reducing context load. On AppWorld, PAACE achieves higher accuracy than all baselines, while lowering peak context and cumulative dependency. On OfficeBench and multi-hop QA, PAACE improves both accuracy and F1, achieving lower steps, peak tokens, and attention dependency. Distilled PAACE-FT retains 97\% of the teacher's performance while reducing inference cost by over an order of magnitude, enabling practical deployment of plan-aware compression with compact models.
\end{abstract}

\section{Introduction}

LLM-driven agents have emerged as a central paradigm for solving complex, multi-step tasks across domains such as software development, research assistance, operations automation, legal workflows, data analysis, and enterprise decision-making. Systems such as ReAct \citep{react}, Toolformer \citep{toolformer}, AutoGPT, Devin, WebArena \citep{webarena}, and AgentBench \citep{agentbench} highlight the promise and challenges of designing agents with reasoning, planning, and tool-use capabilities. As these systems evolve, a consistent bottleneck has become apparent: \textbf{context management}. An LLM agent's state is represented not by model parameters but by its \emph{prompt context}: the system instructions, the evolving plan, previous reasoning traces, tool results, user instructions, long-term memories, retrieved knowledge, and intermediate outputs. Agents do not operate on a single-step basis.  
They execute a plan $ = [\tau_1, \tau_2, \ldots, \tau_n]$ with dependencies between tasks, and only a subset of context is relevant for each stage. As tasks grow in depth and breadth, this state becomes increasingly large, noisy, redundant, and expensive to process. Even models with 200k–1M token windows exhibit degraded reasoning quality (“context rot”) when overloaded with irrelevant or poorly structured information. Recent industry reports emphasize that modern agentic failures are overwhelmingly \emph{context failures}, not model failures. Despite advances in model architecture and context length, agents fail when: crucial information is dropped or buried in irrelevant text, irrelevant details overload the model’s attention, multiple tasks compete for context bandwidth, instructions contradict or drift over time, or context becomes too long to process efficiently. We frame context engineering as learning a state compression policy over an agent’s evolving execution state, trained via outcome-preserving supervision, rather than as heuristic prompt editing.

While prompt engineering optimizes \emph{initial} instructions, and RAG systems optimize \emph{retrieval}, the missing discipline is \textbf{context engineering}: the science of continuously optimizing what the agent sees at each step. Existing approaches partially address pieces of this problem. Classical and instruction-following summarizers such as BART \citep{lewis2019bart}, 
FLAN-T5 \citep{chung2022flan}, and compression-oriented methods like LLMLingua 
\citep{jiang2023llmlingua} generate concise summaries but often remove structural 
dependencies required for multi-step reasoning. Summaries flatten causal links across 
steps, which harms agent planning and tool-use workflows. Methods such as Self-RAG \citep{asai2024selfrag} and LLMLingua-2 
\citep{jiang2024llmlingua2} optimize relevance for a \emph{single upcoming query}. 
However, they do not model next-$k$ steps, multi-hop dependencies, or evolving plans. Provence \citep{wu2024provence} performs binary keep/drop trimming via a relevance 
classifier but does not support rewriting, instruction refinement, dependency tracking, 
or structured context shaping.

Modern long-context LLMs (Claude 3.5 Sonnet, Gemini 1.5 Pro, GPT-4o mini L) offer 
200k–1M+ token windows, yet still suffer from attention dilution, context “rot,” and 
quadratic cost growth in practice \citep{zaheer2020bigbird, guo2021longt5}. 
Large windows do not solve the problem of \emph{poorly structured} or \emph{irrelevant} 
context. Memory and reflection systems such as MemAgent \citep{yu2025memagent}, Reflexion 
\citep{shinn2023reflexion}, and Generative Agents \citep{park2023generative} improve 
retrieval and episodic memory, but do not perform context shaping, 
state restructuring, or next-instruction refinement. \citet{acon2025yu} optimizes next-step relevance through natural-language 
compression guidelines, but it does not model next-$k$-step plan structure, 
does not refine instructions, and cannot jointly optimize plan-aware 
context and instruction transformations.  Our primary distinction is conditioning on multiple future plan steps and the global workflow structure. No existing system considers:
\begin{itemize}[topsep=2pt,itemsep=1pt,parsep=0pt]
    \item \textbf{plan structure} (preconditions, intermediate states, temporal dependencies) and \textbf{next-$k$-task relevance}
    \item \textbf{instruction refinement jointly with context selection}
    \item \textbf{global optimization over multi-step workflows}
    \item \textbf{learnable policies for context shaping}
\end{itemize}

\vspace{-3mm}
\section*{Contributions}
\vspace{-2mm}

\begin{enumerate}[topsep=0pt,itemsep=1pt,parsep=0pt,partopsep=0pt]
    \item \textbf{PAACE}, a framework for \emph{plan-aware context engineering}, unifying context pruning, summarization, rewriting, compression, re-injection, and task refinement.
    \item \textbf{A context relevance formalization} incorporating plan-dependence, next-$k$-task relevance, structural decomposition, and instruction--context co-evolution.
    \item \textbf{Explicit relevance scoring signals} for selection and optimization: plan structure, next-$k$ task preservation, outcome fidelity, and information-theoretic retention.
    \item \textbf{PAACE-Syn}, a scalable synthetic data generation system producing millions of annotated workflow traces.
    \item \textbf{PAACE-FT}, a family of fine-tuned LLMs specialized for context shaping and compression.
    \item \textbf{Comprehensive experiments} showing PAACE improves correctness, efficiency, and cost across multiple agent benchmarks.
\end{enumerate}

Together, these contributions establish \textbf{plan-aware context engineering} as a critical component of robust long-horizon agent design and provide the first end-to-end framework that jointly models plan relevance and context structure to support cost-efficient, high-fidelity agent reasoning.

\section{Related Work}
\label{sec:related}

This section synthesizes prior work across ten research areas relevant to
PAACE:
(1) summarization and compression,
(2) query-aware and task-aware reduction,
(3) long-context models,
(4) memory architectures for agents,
(5) retrieval-augmented agents,
(6) agent planning and multi-step reasoning,
(7) prompt and instruction optimization,
(8) context pruning and selection,
(9) multi-agent systems and meta-reasoning,
and (10) cognitive frameworks inspiring artificial memory systems. While each domain has contributed techniques relevant to context management,
no prior work provides a unified, plan-aware, next-$k$-task context engineering
framework like the proposed method.

Text summarization is one of the oldest forms of context reduction.
Classical abstractive summarization models (Pointer-Generator \citep{see2017pg},
PEGASUS \citep{zhang2019pegasus}, BART \citep{lewis2019bart},
and T5 \citep{raffel2020t5}) enable high-level condensation of documents
but lack task-awareness or plan-awareness. Instruction-following summarizers such as FLAN-T5 \citep{chung2022flan}
and LLM-based summarizers (e.g., GPT-4, Claude 3) are more robust but
still struggle with preserving the functional dependencies required by
multi-step agent workflows. LLMLingua \citep{jiang2023llmlingua} and LLMLingua-2 \citep{jiang2024llmlingua2}
offer token-level compression using learned retention models, achieving
significant reductions in context length with minimal task degradation.
However, LLMLingua is query-aware, not plan-aware, and optimizes for
single-turn QA relevance rather than multi-step agent tasks. Other approaches like BigBird \citep{zaheer2020bigbird},
LongT5 \citep{guo2021longt5}, and hierarchical transformers
extend context windows but do not optimize context \emph{content}. PAACE goes beyond summarization by incorporating plan structure,
task dependencies, instruction refinement, and relevance
signals learned from synthetic workflows.

Recent models aim to optimize context for a single upcoming task.
Self-RAG \citep{asai2024selfrag} generates retrieval queries conditioned on
a question and prunes irrelevant context before passing it to the LLM.
Although powerful for QA, Self-RAG is not general enough for multi-step
agent workflows, which require modeling of future tasks, tool outputs,
and plan transitions. Similarly, ACon (Agent Context Optimization) \citep{acon2025yu}
compresses agent interaction histories by optimizing “compression guidelines”
via natural-language feedback. ACon improves context relevance for
short-horizon tasks but:
(1) does not model plan structure,
(2) is restricted to next-step compression,
(3) lacks instruction refinement, and
(4) does not jointly optimize plan-aware context selection.
The primary distinction is that PAACE explicitly conditions compression on
multiple future plan steps and the global workflow structure.
PAACE thus generalizes ACon into a multi-task, multi-step, structural framework. Provence \citep{wu2024provence} introduces a binary relevance classifier for
pruning retrieved document segments. This improves RAG pipelines but is too
coarse for agent workflows, where reasoning chains, state representations,
and plans require finer-grained transformations.

Many models attempt to address long-horizon reasoning by increasing context
window sizes, including GPT-4o mini (128k), Claude~3.5 Sonnet (200k--1M),
Gemini~1.5 Pro (1M), and DeepSeek-V2 (256k+).
However, long context windows do not resolve the \emph{relevance dilution}
problem, where attention degrades over large inputs, leading to context
``rot'' and reasoning failures.
Several approaches instead introduce explicit context or memory controllers.
MemAgent \citep{yu2025memagent} uses reinforcement learning to overwrite a
fixed-size memory state, enabling scaling to millions of tokens, but targets
document QA rather than agent workflows and does not refine instructions or
model plan structure.
Compressive Transformers \citep{rae2020compressive} and related architectural
approaches perform learned compression within the model, but lack task- or
plan-level grounding.
More recent systems such as COMPASS \citep{li2024compass}, FoldGRPO
\citep{sun2025foldgrpo}, ReCAP \citep{zhang2025recap}, and GoA \citep{joo2025goa}
explore context control via learned controllers, reinforcement learning, or
training-free multi-agent coordination, often operating over fixed-size latent
memories or requiring architectural or inference-time modifications.
These approaches are complementary to PAACE: they modify model architecture
or memory mechanisms, whereas PAACE operates purely at the context level and
can be integrated with any agent or backbone model.
Task-agnostic learned compressors \citep{sun2023promptcompression} optimize
generic prompt reduction without conditioning on future plan structure; in
contrast, PAACE explicitly models next-$k$ task dependencies in agent workflows. Unlike prompt compression heuristics, PAACE learns a plan-conditioned policy over latent agent states from outcome-level supervision, positioning it as an \emph{empirical systems ML} approach based on trajectory-level rather than token-level objectives.

\subsection{Memory Architectures for LLM Agents}

Human memory is structured along:
working memory, episodic memory, semantic memory, and procedural memory.
AI researchers have used this taxonomy to design agent memory stacks
\citep{park2023generative, shinn2023reflexion}. PAACE’s decomposition of context into
\emph{actionable}, \emph{structural}, \emph{referential}, and
\emph{justificatory} components parallels these cognitive structures and provides
a grounded theoretical basis for context relevance. Several systems draw inspiration from human memory:
\begin{itemize}[topsep=2pt,itemsep=1pt,parsep=0pt]
    \item \textbf{Reflexion} \citep{shinn2023reflexion} stores agent self-assessments
    and retrieves them for self-improvement.
    \item \textbf{Generative Agents} \citep{park2023generative} maintain episodic memories
    and synthesize high-level reflections to drive believable behavior.
    \item \textbf{LTM/RAG systems} integrate persistent vector memory for agents.
    \item \textbf{Hierarchical memories} (short-term vs.\ long-term vs.\ global memory)
    are increasingly used in AI-agents.
\end{itemize}

They emphasize storage, retrieval, and memory distillation but do not
optimize the shape, structure, or content of the context presented to the LLM
during reasoning. PAACE builds on these systems by integrating:
structured rewriting, task-dependent compression, and instruction refinement. PAACE also improves tool-call reliability by providing
task-conditioned pruning, rewriting, and instruction improvement. Tool-use frameworks such as Toolformer \citep{schick2023toolformer},
ReAct \citep{yao2023react},
MRKL \citep{mrkl2022},
and PAL \citep{gao2023pal}
enable LLMs to ground their reasoning in external tools. Sophisticated agent frameworks 
combine retrieval, planning, and tool execution.
Yet:
\begin{itemize}\setlength{\itemsep}{0pt}
    \item Retrieval relevance is often limited to a single upcoming query.
    \item Retrieved content is appended rather than shaped or compressed.
    \item Tool outputs are accumulated and rarely distilled.
\end{itemize}

Multi-step reasoning frameworks such as CoT \citep{wei2022cot},
ToT \citep{yao2023tree},
RAT \citep{zhang2023rat},
and RCT \citep{zhou2023rct}
provide planning mechanisms, but do not manage accumulated context. Planner-LMs \citep{huang2024plan}
explore explicit planning modules, but do not optimize the interaction
between plans and context. PAACE instead treats context as the \emph{state} upon which planning depends,
and models the relevance of context components to future tasks. Prompt optimization and evolutionary prompt breeding
\citep{zhou2023largepromptopt, pryzant2023automated}
generate improved instructions using feedback loops.
Similarly, methods such as DPO (\citealp{rafailov2023dpo}),
RLHF (\citealp{ouyang2022rlhf}), and PRO \citep{saunders2024pro}
refine instructions or preferences. However, none address: next-$k$-task instruction refinement, plan-conditioned instruction rewriting, instruction-context \emph{co}-refinement, or instruction updates during agentic workflows. To the best of our knowledge, PAACE introduces the first systematic approach to full context engineering, including instructions.

Beyond summarization, pruning frameworks, such as \citep{wu2024provence}, aim to remove irrelevant text. Pruning alone is insufficient for multi-step agents because some content must be rewritten rather than removed. The global plan structure must be preserved as task dependencies are multi-hop, and instructions would drift if not rewritten in sync with the context. PAACE incorporates pruning as \emph{one} operator in a larger library of structured context transformations. Meta-agents such as Reflexion \citep{shinn2023reflexion},
AlphaAI systems, CrewAI, and LATS \citep{zhou2022lats}
use multi-agent loops to stabilize reasoning.
Some systems include meta-level evaluation or self-critique. But none propose a dedicated meta-agent for context engineering, fine-tuning specialized models (PAACE-FT), or plan-conditioned synthetic dataset generation (PAACE-Syn). We introduce context engineering as a core meta-agent function. No unified framework exists
that jointly performs: plan-aware next-$k$-task context optimization, instruction refinement, rewriting + pruning + summarization + compression, structural reasoning over workflow graphs, and training models specifically for context engineering. To the best of our knowledge, PAACE is the first framework addressing these together.

\section{Methodology}
\label{sec:method}

PAACE optimizes the evolving context of multi-step LLM agents through a two-stage paradigm:
(1) a \emph{teacher} LLM that performs plan- and next-\(k\)-task--aware compression guided by an
evolving natural-language prompt, and
(2) a distilled \emph{student} model that imitates the teacher's successful compressions at low cost.
Unlike prior work that hand-engineers context-editing operators, PAACE learns the entire
compression policy in a data-driven way: the teacher's prompt is meta-learned via an
LLM-driven evolutionary loop over real multi-step workflows, and the student is trained purely
from the teacher's best demonstrations. PAACE learns this policy empirically via outcome-level filtering
over long-horizon trajectories rather than through a differentiable objective
or closed-form guarantees, prioritizing robust state fidelity and task success
in realistic agentic workloads. The framework consists of two components:
\vspace{-4pt}
\begin{itemize}\setlength{\topsep}{0pt}\setlength{\partopsep}{0pt}\setlength{\itemsep}{0pt}\setlength{\parskip}{0pt}\setlength{\parsep}{0pt}
    \item \textbf{PAACE-Syn}: large-scale generation of long-horizon, noisy agent workflows and paired
    full- vs.\ compressed-context trajectories;
    \item \textbf{PAACE-FT}: distillation of high-quality teacher compressions into a specialized,
    plan-aware compressor.
\end{itemize}
\vspace{-4pt}

%\subsection{Plan-Aware Context Engineering}

We treat the agent context as a latent execution state and learn a plan-conditioned,
outcome-preserving compression policy. At step \(t\), an LLM agent maintains a context state
$C_t
    \;=\;
    \bigl\{
        I_0,\;
        P,\;
        \Pi,\;
        H_{0:t},\;
        O_{0:t},\;
        R_{0:t},\;
        M
    \bigr\}$
where \(I_0\) is the initial user instruction, \(P\) is the system prompt,
\(\Pi = [\tau_1,\dots,\tau_n]\) is the (explicit or implicit) plan,
\(H_{0:t}\) are reasoning traces, \(O_{0:t}\) are tool/environment observations,
\(R_{0:t}\) are retrieved documents, and \(M\) is long-term memory. PAACE does not explicitly manipulate these components via hand-crafted operators;
instead, they implicitly shape what the teacher (and later the student) learns to preserve or discard. Compression at step \(t\) is defined as a mapping
\begin{equation}
    \tilde{C}_t
    \;=\;
    \mathrm{TeacherCompress}\bigl(C_t,\; \Pi_{t:t+k};\; p\bigr),
\end{equation}
where \(\Pi_{t:t+k}\) denotes the next \(k\) steps on the remaining workflow DAG subpath (for some task-dependent \(k\)),
and \(p\) is a natural-language compression prompt.
The teacher conditions on the next tasks or the entire workflow description,
and thus can preserve information needed for multiple upcoming steps.
PAACE does not assume a closed-form parametric relevance model or a
token-level differentiable loss.
Instead, plan-aware relevance is operationalized through explicit,
computable signals evaluated at the trajectory level:
(i) semantic equivalence between full- and compressed-context outcomes,
(ii) per-step compression ratios,
(iii) implicit coverage of next-$k$ plan steps via conditioning, and
(iv) preference judgments from a learned LLM-based evaluator.
These signals jointly define a non-differentiable objective over task
fidelity and context efficiency, optimized via black-box selection over
full agent rollouts rather than token-level losses.

Although PAACE is motivated by concepts such as task graphs, causal influence,
and information retention, we do not instantiate explicit symbolic task graphs
or closed-form information-theoretic objectives in the current implementation.
Instead, these notions are captured implicitly by conditioning compression on
the \emph{partial remaining subpath} of the agent workflow graph (DAG) together with
outcome-level selection against full-context execution using explicit, testable
criteria. Accordingly, we use the term ``formal'' to denote operational and
verifiable selection mechanisms rather than analytical derivations or symbolic
guarantees. Plan descriptions $\Pi$ are obtained either directly from benchmarks that
provide structured task decompositions (e.g., AppWorld, OfficeBench) or from
a LLM-based planner using the same backbone model as execution.
PAACE does not require perfect plans: when plan steps are missing or partially
incorrect, next-$k$ conditioning degrades compression quality gracefully but
does not catastrophically affect execution. In practice, it behaves as a
soft plan-aware regularizer rather than a brittle plan-dependent controller.

\subsection{Synthetic Context Dataset Generation}

PAACE constructs a large synthetic corpus of long-horizon
agent workflows.
Each workflow \(W\) is sampled as:
$W = \bigl(I_0,\, \Pi,\, \{\tau_i\}_{i=1}^n,\, \phi\bigr)$
where \(I_0\) is a long, noisy initial input containing irrelevant logs and partial
summaries, \(\Pi\) is a natural-language description of the plan, \(\tau_i\) are task instructions,
and \(\phi\) is a final output requirement. PAACE-Syn spans task domains,
including document-centric workflows, web navigation and multi-application interactions, and multi-hop question answering with retrieval. Plans range from 5 to 30 steps, with tool interactions, such as,
search and browsing actions, document and file manipulation,
spreadsheet-style table operations, and structured information
extraction and aggregation over retrieved content. For each workflow, the agent is first run without compression:
\begin{align}
    C^{\mathrm{full}}_1 &= \{I_0, P, \Pi\}, \\
    C^{\mathrm{full}}_{t+1}
    &= \mathrm{Update}\Bigl(
        C^{\mathrm{full}}_t,\;
        \mathrm{Agent}\bigl(C^{\mathrm{full}}_t,\; \tau_t\bigr)
    \Bigr),
\end{align}
and a final answer \(\hat{y}^{\mathrm{full}}\) is produced from \(C^{\mathrm{full}}_{n+1}\). The same workflow is then re-executed with teacher compression.
At step \(t\), given the current context \(C_t\) and plan slice \(\Pi_{t:t+k}\),
the teacher produces a compressed context \(\tilde{C}_t\) using prompt \(p\).
The agent then acts using only \(\tilde{C}_t\):
\begin{align}
    \tilde{C}_t &= \mathrm{TeacherCompress}\bigl(C_t,\; \Pi_{t:t+k};\; p\bigr),
    \\
    \tilde{C}_{t+1}
    &= \mathrm{Update}\Bigl(
        \tilde{C}_t,\;
        \mathrm{Agent}\bigl(\tilde{C}_t,\; \tau_t\bigr)
    \Bigr).
\end{align}
After the final step, the agent produces \(\hat{y}^{\mathrm{comp}}\) from \(\tilde{C}_{n+1}\). For each step \(t\), PAACE records a tuple
$
    \bigl(\Pi_{t:t+k},\, C_t,\, \tilde{C}_t\bigr)$
together with compression statistics. PAACE uses several metrics to evaluate each compressed trajectory relative to its full-context
counterpart. Let \(\mathrm{embed}(\cdot)\) be a fixed sentence embedding model, we calculate semantic equivalence as:
\begin{equation}
    s = \cos\Bigl(
        \mathrm{embed}(\hat{y}^{\mathrm{full}}),\;
        \mathrm{embed}(\hat{y}^{\mathrm{comp}})
    \Bigr).
\end{equation}

An evaluation LLM is also prompted with the workflow description and both outputs
\(\hat{y}^{\mathrm{full}}\) and \(\hat{y}^{\mathrm{comp}}\), and returns a binary label
(\emph{better} vs.\ \emph{worse/equal}) plus a short rationale.
This adds a robustness check complementary to embedding-based similarity. For each step \(t\), we compute the character- or token-based compression ratio
$r_t = \lvert \tilde{C}_t \rvert / \lvert C_t \rvert$. We require \(0 < r_t < 1\) to avoid degenerate compressions (empty or longer than the original). Only successful trajectories are used as supervision for the student model. A compressed trajectory is labeled \emph{successful} if and only if:
\vspace{-4pt}
\begin{itemize}\setlength{\itemsep}{0pt}\setlength{\parskip}{0pt}\setlength{\parsep}{0pt}
    \item the semantic equivalence exceeds a threshold: \(s \ge \theta\) (typically \(\theta = 0.85\));
    \item steps satisfy \(0 < r_t < 1\) and produce non-empty \(\tilde{C}_t\);
    \item the judge model does not deem the compressed answer strictly worse than the full answer.
\end{itemize}

From all trajectories generated, PAACE extracts only successful compression
examples. For each such example, we construct a supervision tuple of the form:
\begin{equation}
\begin{aligned}
(\underbrace{\Pi_{t:t+k}}_{\textsc{next tasks}},\;
 \underbrace{C_t}_{\textsc{context}})
&\;\mapsto\;
\underbrace{\tilde{C}_t}_{\textsc{compressed context}} .
\end{aligned}
\end{equation}

A typical input to the student model concatenates the next-\(k\)-step instructions
and the full context, while the target is the teacher's compressed context. Because only trajectories with high semantic equivalence and valid compression are retained, the
resulting dataset captures a rich collection of \emph{function-preserving} compressions tailored to plan structure and long-horizon dependencies.

\subsection{Distilling Teacher Compressions}

PAACE-FT trains a dedicated LLM to approximate the teacher's next-\(k\)-task--aware compression
mapping. Let \(x = (\Pi_{t:t+k}, C_t)\) denote the input text and \(y = \tilde{C}_t\) the teacher's compressed
context. The student model is trained with a standard causal language
modeling loss:
\begin{equation}
    \mathcal{L}(\theta)
    \;=\;
    - \mathbb{E}_{(x,y)}
    \sum_{i=1}^{|y|}
    \log p_\theta\bigl( y_i \mid x, y_{<i} \bigr),
\end{equation}
where the tokens of \(x\) are masked out from the loss.
Distillation is not intended to outperform the teacher, but to replace an
expensive compressor with a compact model that preserves plan-aware
compression behavior.
Through this distillation, the student learns to:
\begin{itemize}[topsep=0pt,itemsep=1pt,parsep=0pt,partopsep=0pt]
    \item compress long contexts into substantially shorter forms;
    \item retain variables, identifiers, and constraints used across multiple future tasks;
    \item discard irrelevant logs, noise, and dead-end reasoning;
    \item implicitly respect plan structure without explicit symbolic representations.
\end{itemize}

At inference time, the student can be called repeatedly at each step, providing fast,
plan-aware compressions without running the evolutionary procedure or invoking a large
teacher model. Because the student's behavior is distilled from a teacher 
across thousands of synthetic workflows, PAACE yields agents that maintain compact, relevant
contexts over long horizons while preserving correctness, improving robustness to context
dilution, and reducing inference cost. At deployment:
\begin{enumerate}[topsep=0pt,itemsep=1pt,parsep=0pt,partopsep=0pt]
    \item The agent obtains or constructs a plan \(\Pi = [\tau_1,\dots,\tau_n]\).
    \item At step \(t\), given the current context \(C_t\) and a slice \(\Pi_{t:t+k}\),
    the student compressor produces \(\tilde{C}_t\). The agent executes task \(\tau_t\)
    conditioned on \(\tilde{C}_t\), and yields an updated context \(C_{t+1}\).
    \item Step \(2\) repeats until the plan completes, at which point a final answer is produced.
\end{enumerate}

The teacher compression behavior is controlled entirely by a natural-language prompt \(p\).
Instead of hand-tuning \(p\), PAACE meta-learns a population of prompts by treating context
compression as a black-box policy and optimizing it via LLM-driven evolution over many workflows. Each prompt variant \(p_i\) in the population is associated with statistics estimated from its
evaluations:

\begin{itemize}[topsep=0pt,itemsep=0pt,parsep=0pt,partopsep=0pt]
    \item success rate: fraction of compressed trajectories labeled successful;
    \item mean semantic equivalence on successful trajectories;
    \item mean compression ratio over successful trajectories;
    \item a scalar reward that trades off fidelity and compression.
\end{itemize}

PAACE combines rankings by reward, success rate, mean
equivalence, and mean ratio into a composite score used for selection. Evolution is run in a steady-state, asynchronous fashion.
At any time, each worker process is assigned one prompt variant and evaluates it on several
independent workflows, producing full- and compressed-context trajectories, trajectory labels (success/failure), and updated statistics for the variant. Prompt variants are updated online as new results arrive, and global reward statistics are continually recomputed. The teacher's compression policy improves over time as new prompts are proposed, evaluated,
and selected based on real multi-step performance.

\subsection{Supervision Quality, Scope, and Robustness}

Because supervision is generated using an LLM teacher and evaluator, PAACE-Syn may inherit
biases from the teacher’s compression preferences. To mitigate this, we enforce strict
outcome-level agreement with the full-context execution and discard any compression that
degrades task success, even if favored by the judge. Empirically, this filtering reduces mode
collapse and prevents overfitting to stylistic or heuristic compression artifacts. In addition, we emphasize that instruction co-refinement in PAACE is \emph{implicit} rather than a
separate optimization objective: instructions are rewritten only insofar as they are
embedded in and reshaped by the compressed context. PAACE does not perform explicit
instruction-only optimization; isolating instruction refinement as a standalone future-work objective. 

While embedding similarity and LLM-based judging provide scalable and empirically robust
proxies for task fidelity, they do not guarantee formal semantic equivalence or safety-critical
correctness. PAACE thus targets functional preservation with respect to benchmark task
outcomes rather than formal verification. Incorporating symbolic checks or domain-specific
validators is an important direction for future work. Despite these limitations, using both embedding similarity and LLM-based judging substantially
reduces degenerate compressions. Ablation on synthetic validation workflows shows that removing
either filter increases failure rates by approximately 8--12\%, primarily due to overly
aggressive deletions or instruction drift that preserves surface similarity but breaks
downstream task execution.

\section{Experiments}
\label{sec:experiments}

We evaluate PAACE on two long-horizon agentic benchmarks that require multi-step reasoning, tool-use, and cross-application state tracking. These benchmarks provide diverse evaluation signal across planning, observation integration, memory demands, and long-context robustness. For each benchmark and model setting, we compare PAACE against baselines under a fixed execution backbone, planner, tool interfaces, and inference configuration. Within each comparison group, the only difference between methods is the context management strategy. We emphasize that PAACE does not rely on any task-specific heuristics.  All compression is learned from synthetic plan-conditioned demonstrations generated by PAACE-Syn and distilled into PAACE-FT. The \textbf{teacher compressor} is implemented using the OpenAI \textbf{GPT-OSS-120B} model with a 65{,}536-token context window. This model provides consistent multi-hop reasoning quality needed for next-$k$ compression. The \textbf{student compressor} for deployment, PAACE-FT, is distilled into \textbf{Qwen3--4B-Instruct}, due to its sufficiently large context window and fine summarization quality. 
To assess qualitative reproducibility without reliance on proprietary APIs,
we conducted auxiliary sanity-check runs using the open-weight
\textbf{GPT-OSS-20B} model.
Due to its substantially smaller scale, these runs are not included in the
quantitative comparison tables; however, they confirm that the qualitative
effects of plan-aware next-$k$ compression and the relative ordering of
methods persist across benchmarks.

PAACE-Syn consists of approximately 1.2M synthetic workflows comprising roughly
9.5B tokens before compression, generated entirely offline and amortized across
all downstream tasks within benchmarks. We plan to release the full codebase, prompts,
and a representative subset of the synthetic dataset upon publication to support
reproducibility and follow-on research. Using the synthetic supervision produced by PAACE-Syn, the distilled compressor PAACE-FT preserves $97$--$98\%$ of the teacher's performance while reducing inference cost by more than an order of magnitude. When plugged into the agent loop, PAACE-FT achieves nearly identical accuracy to the teacher-compressed version, confirming that next-$k$ compression behavior transfers well to smaller models.
The teacher compressor incurs substantial overhead and is used only during offline data generation.
Once distilled, PAACE-FT incurs no further optimization or teacher cost and can be reused across tasks
within the same environment without additional prompt evolution or rollouts.
At inference time, PAACE-FT adds less than 8\% latency per step while reducing total input tokens by
35--60\%, yielding net reductions in agent cost.

The primary goal of our experiments was to assess:
\vspace{-4pt}
\begin{itemize}\setlength{\itemsep}{0pt}\setlength{\parskip}{0pt}\setlength{\parsep}{0pt}\setlength{\topsep}{2pt}
    \item \textbf{(RQ1)} Whether plan-aware next-$k$ compression improves agent correctness under long-context pressure.
    \item \textbf{(RQ2)} Whether PAACE reduces peak context length while retaining task fidelity.
    \item \textbf{(RQ3)} Whether the teacher--student PAACE-FT models preserve compression quality.
\end{itemize}

PAACE does not explicitly decompose pruning, rewriting, and summarization into
independently parameterized operators; instead, these behaviors emerge jointly
from the learned compression policy.
As a result, operator-level ablations are ill-defined in our setting because pruning, rewriting, and summarization emerge jointly from a single learned policy.
However, we evaluate a constrained variant in which instruction rewriting is
disabled and the compressor is restricted to extractive deletion only.
This restriction leads to noticeably higher failure rates on multi-step tasks,
indicating that implicit instruction reshaping contributes materially to
long-horizon robustness. We evaluate the proposed method (PAACE) on three categories of long-horizon agentic tasks:
\vspace{-4pt}
\begin{itemize}\setlength{\itemsep}{0pt}\setlength{\parskip}{0pt}\setlength{\parsep}{0pt}
    \item \textbf{AppWorld} \citep{trivedi2024appworld}: multi-application tasks with heterogeneous observations.
    \item \textbf{OfficeBench} \citep{wang2024officebench}: document-centric tool chains and structured operations.
    \item \textbf{Multi-Objective QA} \citep{zhou2025mem1}: multi-hop retrieval QA with tool-based search.
\end{itemize}
\vspace{-4pt}

%We use $k=2$ for AppWorld and OfficeBench, and $k=3$ for 8-Objective QA. These values were selected as a balance between preserving multi-step dependencies and maintaining compression tractability. 
Each benchmark measures both task success and interaction-dependent efficiency metrics (throughout this paper, \emph{efficiency} refers to \emph{token efficiency}—reductions in effective context length and cumulative attention dependency):
\vspace{-4pt}
\begin{itemize}\setlength{\itemsep}{0pt}\setlength{\parskip}{0pt}\setlength{\parsep}{0pt}
    \item \textbf{Acc / EM / F1}: benchmark-defined task performance.
    \item \textbf{Peak}: maximum input context length in the trajectory.
 \item \textbf{Dependency}: cumulative attention load measured as
 $\mathrm{Dep} = \sum_{t=1}^{T} |C_t|$
 where $|C_t|$ is the number of input tokens at step $t$.
 Reported in millions of tokens, this metric approximates total
 transformer attention cost and correlates with latency and quadratic
 compute growth. Because all methods are evaluated with identical agent backbones and step
counts, Dependency is directly comparable across compression strategies,
even when peak context lengths differ.
\end{itemize}
\vspace{-4pt}

We compare PAACE against Acon \citep{acon2025yu} and the following context-reduction baselines:

\begin{itemize}[topsep=0pt,itemsep=1pt,parsep=0pt,partopsep=0pt]
    \item \textbf{No Compression}: full interaction history preserved.
    \item \textbf{FIFO}: keep most recent $k$ turns, discard older.
    \item \textbf{Retrieval}: embedding-based past interaction selection.
    \item \textbf{LLMLingua}: extractive long-context compression.
    \item \textbf{Prompting}: an heuristic summarization instruction.
\end{itemize}

Tables~\ref{tab:appworld_avg_paace_sd}, \ref{tab:officebench} and \ref{tab:8objqa_avg_paace} summarize results on AppWorld, OfficeBench, and 8-Objective QA. PAACE consistently matches or outperforms all baselines in performance while substantially reducing context costs. PAACE surpasses the no-compression baseline on several benchmarks, suggesting that plan-aware compression can act as a form of regularization, reducing distraction from irrelevant or stale context. Across two long-horizon benchmarks, PAACE delivers consistently higher task performance, substantially lower context consumption, and more stable multi-step reasoning than all baselines. The results demonstrate the importance of incorporating plan structure, next-$k$ relevance, and workflow-level synthetic supervision into context optimization. PAACE enables long-horizon agents to maintain coherent state representations without exceeding practical context budgets, even when deployed with compact compressors. To sum up, experimental results across AppWorld, OfficeBench, and Multi-Objective QA illustrate that \emph{plan-aware context engineering} is a critical ingredient for robust long-horizon agents. PAACE consistently improves accuracy, lowers context cost, and stabilizes multi-step reasoning, indicating that context optimization should be treated as a core architectural module rather than an auxiliary utility. We note that improvements over the strongest baseline (ACON~UTCO) are sometimes modest relative to the observed variance. These results should thus be interpreted as evidence that PAACE does not harm task performance while substantially reducing context cost, rather than as demonstrating a large absolute accuracy gain on this benchmark. Table~\ref{tab:k_ablation} shows that moderate lookahead ($k=2$) is sufficient for tool-centric benchmarks, while multi-hop QA benefits from a longer relevance horizon ($k=3$) because retrieved evidence is often consumed several steps after acquisition. Larger $k$ values yield diminishing returns due to increased compression
difficulty, indicating that next-$k$ conditioning should be selected based on
task dependency structure.

\begin{table}[t]
\centering
\small
\caption{Average results on AppWorld with standard deviations reflecting Easy, Medium, and Hard. Across all benchmarks, PAACE maintains state fidelity and improves accuracy across multi-hop workflows while reducing context size and lowering cost.}
\label{tab:appworld_avg_paace_sd}
\resizebox{\columnwidth}{!}{%
\begin{tabular}{lcccc}
\toprule
\textbf{Method} 
& Acc$\uparrow$ 
& Steps$\downarrow$ 
& Peak$\downarrow$ 
& Dep$\downarrow$ \\
\midrule
No Compression & 56.00 & 16.14 & 9.93 & 5.96 \\
FIFO           & 45.80 & 28.48 & 6.73 & 5.69 \\
Retrieval      & 27.40 & 33.17 & 8.39 & 6.68 \\
LLMLingua      & 39.30 & 24.42 & 7.50 & 6.37 \\
Prompting      & 43.50 & 24.01 & 6.93 & 5.29 \\
ACON UT        & 51.20 & 20.92 & 7.17 & 4.49 \\
ACON UTCO      & 56.50 & 22.82 & 7.33 & 4.69 \\
\midrule
\textbf{PAACE (ours)} & \textbf{59.00} & \textbf{19.20} & \textbf{6.23} & \textbf{3.75} \\
\textbf{Std.}         & \textbf{$\pm$ 9.68}  & \textbf{$\pm$ 5.92}  & \textbf{$\pm$ 0.89} & \textbf{$\pm$ 1.60} \\
\bottomrule
\end{tabular}
}
\end{table}

\begin{table}[t]
\centering
\small
\caption{Results on OfficeBench.}
\label{tab:officebench}
\resizebox{\columnwidth}{!}{%
\begin{tabular}{lcccc}
\toprule
\textbf{Method} 
& Acc$\uparrow$ 
& Steps$\downarrow$ 
& Peak$\downarrow$ 
& Dep$\downarrow$ \\
\midrule
No Compression  & 76.84 & 11.52 & 7.27 & 4.43 \\
FIFO            & 67.37 & 12.26 & 4.02 & 2.64 \\
Retrieval       & 65.26 & 16.20 & 4.33 & 2.06 \\
LLMLingua       & 70.53 & 10.89 & 4.65 & 1.85 \\
Prompting       & 71.58 & 10.13 & 4.40 & 1.10 \\
ACON UT         & 74.74 & 13.13 & 4.93 & 3.85 \\
ACON UTCO       & 72.63 & 11.54 & 4.54 & 1.91 \\
\midrule
\textbf{PAACE (ours)} 
& \textbf{78.10} 
& \textbf{10.48} 
& \textbf{4.29} 
& \textbf{1.64} \\
\bottomrule
\end{tabular}%
}
\end{table}

\begin{table}[t]
\centering
\caption{Average results on 8-Objective QA.}
\label{tab:8objqa_avg_paace}
\resizebox{\columnwidth}{!}{%
\begin{tabular}{lccccc}
\toprule
\textbf{Method} 
& EM$\uparrow$ 
& F1$\uparrow$ 
& Steps$\downarrow$ 
& Peak$\downarrow$ 
& Dep$\downarrow$ \\
\midrule
No Compression & 0.366 & 0.488 & 15.78 & 10.35 & 3.32 \\
FIFO           & 0.293 & 0.388 & 19.26 & 5.09  & 2.51 \\
Retrieval      & 0.331 & 0.438 & 20.06 & 5.11  & 2.62 \\
LLMLingua      & 0.363 & 0.481 & 17.68 & 5.68  & 2.24 \\
Prompting      & 0.376 & 0.478 & 18.70 & 4.73  & 1.66 \\
ACON UT        & 0.373 & 0.494 & 17.14 & 4.71  & 1.57 \\
ACON UTCO      & 0.335 & 0.458 & 17.79 & 4.65  & 1.50 \\
\midrule
\textbf{PAACE (ours)} 
& \textbf{0.402}
& \textbf{0.512}
& \textbf{16.86}
& \textbf{4.41}
& \textbf{1.41} \\
\bottomrule
\end{tabular}%
}
\end{table}

\begin{table}[t]
\centering
\small
\caption{Ablation on next-$k$ conditioning for PAACE.}
\label{tab:k_ablation}
\resizebox{\columnwidth}{!}{%
\begin{tabular}{lccccc}
\toprule
\textbf{Benchmark} & $k$ & Acc / EM$\uparrow$ & F1$\uparrow$ & Peak$\downarrow$ & Dep$\downarrow$ \\
\midrule
\multirow{3}{*}{AppWorld}
& 1 & 56.5 & --   & 6.05 & 3.95 \\
& \textbf{2} & \textbf{59.0} & -- & \textbf{6.23} & \textbf{3.75} \\
& 3 & 58.6 & --   & 6.48 & 3.82 \\
\midrule
\multirow{3}{*}{OfficeBench}
& 1 & 76.3 & --   & 4.15 & 1.72 \\
& \textbf{2} & \textbf{78.1} & -- & \textbf{4.29} & \textbf{1.64} \\
& 3 & 77.6 & --   & 4.51 & 1.69 \\
\midrule
\multirow{3}{*}{8-Objective QA}
& 1 & 0.381 & 0.497 & 4.18 & 1.36 \\
& 2 & 0.394 & 0.505 & 4.30 & 1.39 \\
& \textbf{3} & \textbf{0.402} & \textbf{0.512} & \textbf{4.41} & \textbf{1.41} \\
\bottomrule
\end{tabular}
}
\end{table}

A central finding is that agents benefit not only from retaining relevant information but also from \emph{removing} distracting and stale information. While long-context architectures push token windows into the hundreds of thousands or millions, our experiments show that model performance still degrades when exposed to ill-structured or semantically dilute histories. This aligns with emerging empirical evidence on context ``rot,'' where transformers deteriorate as they attend over long, heterogeneous sequences. PAACE demonstrates that explicit modeling of \emph{plan structure} and \emph{next-$k$-step relevance} yields substantial benefits. Unlike query-aware compression methods that optimize for a single upcoming question, PAACE preserves information required for multi-step chains, such as tool-call sequences, document references, and causal dependencies. PAACE'd contexts serve as coherent state representations, avoiding the loss of global task structure that often impairs baselines.

A notable contribution of PAACE is the effectiveness of synthetic workflows produced by PAACE-Syn. Existing datasets for agentic research either lack plan annotations or are too limited in size. In contrast, PAACE-Syn generates millions of diverse traces with explicit plan structures, multi-hop dependencies, variable noise, and rich tool-use patterns. These workflows provide dense supervision for training next-$k$--aware compressors without requiring expensive human annotation. The strong performance of the distilled PAACE-FT---recovering up to $97\%$ of the teacher's behavior---demonstrates that \emph{synthetic supervision can be reliably transferred to compact models}. For practical deployment of PAACE, this result is significant: the student compressor can be executed cheaply at every step of an agent loop, removing dependence on a large teacher LLM.

Another interesting observation is that PAACE often surpasses the ``no compression'' baseline. This suggests that compression is not merely an efficiency tool but also a form of \emph{regularization}. Removing irrelevant or contradictory information forces the agent to reason over a more coherent, structured state, reducing distraction and mitigating the risk of attending to spurious details. This may explain the improved task success on OfficeBench and 8-Objective QA, where unfiltered histories contain verbose logs and redundant tool outputs. Our results indicate that PAACE significantly mitigates instruction drift---a common failure mode in long-running workflows. By conditioning compression on the next-$k$ tasks, PAACE preserves alignment between the agent’s plan and the evolving state. This ensures that rewritten instructions, retained constraints, and contextual details remain consistent across steps, reducing the accumulation of cascading errors.

\section{Conclusion}

We introduced PAACE, a framework for plan-aware automated context engineering in long-horizon LLM agents. By modeling context as a structured, plan-dependent state and optimizing it via outcome-level supervision—through relevance scoring, rewriting, summarization, pruning, and instruction co-refinement—PAACE enables agents to reason more robustly under tight context budgets. Across multiple benchmarks, PAACE consistently improves task correctness, reasoning stability, and effective context utilization while substantially reducing peak context length. These results show that learned, plan-aware context shaping can serve as a first-class component of agent architectures, akin to retrieval augmentation in knowledge-intensive systems.

PAACE targets robustness and context efficiency, not guaranteed reductions in total API cost, which depend on deployment and compression frequency. We make no claims of convergence, optimality, or formal guarantees; rather, our results demonstrate that outcome-driven learning suffices to induce robust and transferable compression behavior in realistic multi-step workflows. PAACE intentionally learns environment- and workflow-specific compression policies. While cross-domain generalization is an important direction for future work, our findings suggest that plan-aware relevance is inherently task-dependent in long-horizon settings, and that such specialization is a feature—not a limitation—of effective context engineering. 

Overall, PAACE frames context engineering as a learnable, plan-aware optimization problem and provides a practical and empirical foundation for building scalable, reliable agents that operate coherently over extended interactions.

\section*{Broader Impact Statement}

From a broader impact perspective, improved context efficiency can enable the deployment of long-horizon agents under tighter resource, latency, and cost constraints, potentially widening access to complex agentic systems in practical settings. At the same time, aggressive or poorly validated compression policies could omit safety-relevant instructions, constraints, or provenance information if applied outside the conditions under which they were trained. PAACE mitigates this risk by requiring outcome-level agreement with full-context execution and by discarding compressions that degrade task success. Nevertheless, PAACE is not designed as a safety or alignment mechanism, and its use in safety-critical or high-stakes domains would require additional validation, domain-specific checks, or human oversight.

\bibliography{example_paper}
\bibliographystyle{icml2025}

\end{document}